\pdfoutput=1

\documentclass[11pt]{article}

\usepackage{acl}

\usepackage{times}
\usepackage{latexsym}

\usepackage[T1]{fontenc}

\usepackage[utf8]{inputenc}

\usepackage{microtype}

\usepackage{tikz}
\usetikzlibrary{decorations.pathreplacing}
\usepackage{subcaption}
\usepackage{graphicx}

\usepackage{booktabs}
\usepackage{multirow}

\usepackage{amsmath}
\usepackage{amsfonts}

\newcommand{\defn}[1]{\textbf{#1}}

\DeclareMathOperator*{\argmax}{argmax}

\newcommand{\sep}{\textsc{sep}}
\newcommand{\eos}{\textsc{eos}}

\newcommand{\cls}[1]{\textsc{cls}[#1]}

\newcommand{\R}{\mathbb{R}}

\newcommand{\xx}{\boldsymbol{x}}
\newcommand{\s}{\boldsymbol{s}}
\newcommand{\pp}{\boldsymbol{p}}
\newcommand{\ww}{\boldsymbol{w}}
\newcommand{\qq}{\mathbf{q}}
\newcommand{\kk}{\mathbf{k}}
\newcommand{\vv}{\mathbf{v}}
\newcommand{\zz}{\mathbf{z}}

\newcommand{\aaa}{\mathbf{a}}

\newcommand{\softmax}{\mathrm{softmax}}

\everypar{\looseness=-1}

\usepackage{xcolor}
\definecolor{color1}{RGB}{102,194,165}
\definecolor{color2}{RGB}{141,160,203}
\definecolor{color3}{RGB}{252,141,98}
\definecolor{light-gray}{gray}{0.8}

\usepackage{cleveref}
\crefname{section}{\S}{\S\S}
\Crefname{section}{\S}{\S\S}
\crefname{table}{Tab.}{}
\crefname{figure}{Fig.}{}
\crefname{algorithm}{Alg.}{}
\crefname{appendix}{App.}{}
\crefname{lemma}{Lemma}{}
\Crefname{theorem}{Theorem}{}
\crefname{prop}{Proposition}{}
\crefname{cor}{Corollary}{}
\crefname{align}{}{}
\crefname{equation}{}{}

\usepackage[disable]{todonotes}
\usepackage{inconsolata}

\title{Probing via Prompting}

\author{Jiaoda Li~\;~Ryan Cotterell~\;~Mrinmaya Sachan
\\
   {%
\setlength{\fboxsep}{2.5pt}%
\setlength{\fboxrule}{2.5pt}%

\fcolorbox{white}{white}{
  $\{${\href{mailto:jiaoda.li@inf.ethz.ch}{\tt jiaoda.li}, }{\href{mailto:ryan.cotterell@inf.ethz.ch}{\tt ryan.cotterell}, } \href{mailto:mrinmaya.sachan@inf.ethz.ch}{\tt mrinmaya.sachan}$\}${\tt @inf.ethz.ch}}} \\
    {%
\setlength{\fboxsep}{2.5pt}%
\setlength{\fboxrule}{2.5pt}%

\fcolorbox{white}{white}{\includegraphics[width=.15\linewidth]{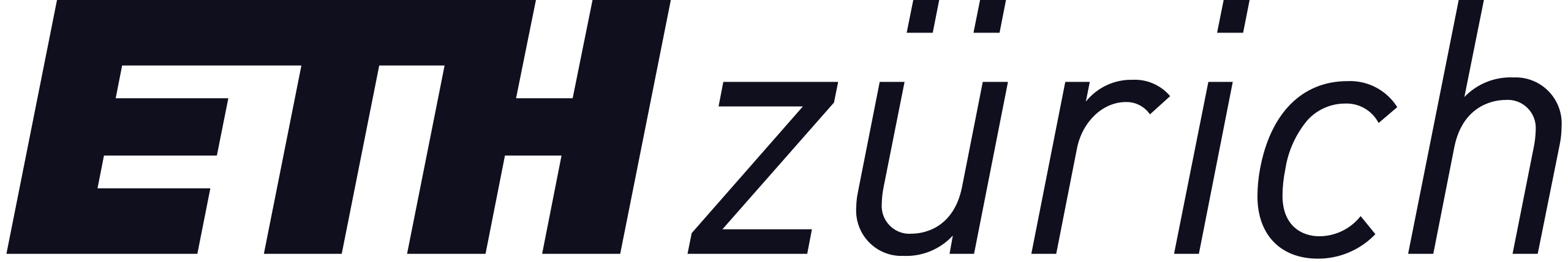}}}
}

\begin{document}
\maketitle
\begin{abstract}
Probing is a popular method to discern what linguistic information is contained in the representations of pre-trained language models.
However, the mechanism of selecting the probe model has recently been subject to intense debate, as it is not clear if the probes are merely extracting information or modeling the linguistic property themselves.
To address this challenge, this paper introduces a novel model-free approach to probing, by formulating probing as a prompting task.
We conduct experiments on five probing tasks and show that our approach is comparable or better at extracting information than diagnostic probes while learning much less on its own. 
We further combine the probing via prompting approach with attention head pruning to analyze where the model stores the linguistic information in its architecture. 
We then examine the usefulness of a specific linguistic property for pre-training by removing the heads that are essential to that property and evaluating the resulting model's performance on language modeling.
\newline
\newline
\vspace{1.5em}
\hspace{.5em}\includegraphics[width=1.25em,height=1.25em]{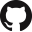}\hspace{.75em}\parbox{\dimexpr\linewidth-2\fboxsep-2\fboxrule}{\url{https://github.com/rycolab/probing-via-prompting}}
\vspace{-.5em}
\end{abstract}

\section{Introduction}

Pre-trained language models such as BERT \cite{Devlin:2018}, GPT-2 \cite{radford2019language}, and GPT-3 \cite{Larochelle2020} have increased the performance of data-driven natural language processing (NLP) models on a wide variety of tasks.
Due to their strong performance on many language-based tasks that require some linguistic understanding, it is natural to hypothesize that the models must implicitly encode some linguistic knowledge.
One avenue of research that attempts to uncover the linguistic knowledge encoded in these models is called probing \cite{conneau-etal-2018-cram, alain2018understanding, tenney2018what, saleh-etal-2020-probing}.
A common form of probing is diagnostic probing. Under this approach a classifier is trained on top of a pre-trained language model to perform a target linguistic task, which is closely related to the linguistic property in question.
The predictive accuracy of the classifier is then taken as an indicator of how much knowledge about the target linguistic property is encoded in the language model representations.\looseness=-1

However, diagnostic probing has its limitations. An inherent challenge in the endeavor is discerning what is encoded in the pre-trained representations from what is learned by the probe itself \cite{zhang-bowman-2018-language, hewitt-liang-2019-designing,pimentel-etal-2020-pareto,cao-etal-2021-low}.
The probe could, in principle, learn the task on top of random representations.
While the probe is trained to extract linguistic properties from the representations, there is no simple way to restrain the probe from learning the task on its own during training.
Previous research tackles the challenge
using random model baselines \cite{zhang-bowman-2018-language} and control tasks
\cite{hewitt-liang-2019-designing, pimentel-etal-2020-pareto}. \citet{cao-etal-2021-low} try to create a more selective probe using pruning.\looseness=-1

\begin{figure*}
\footnotesize
    \centering
    \begin{subfigure}[t]{0.4\textwidth}
        \centering
        \begin{tikzpicture}[
            text_node/.style={draw, minimum width=0.8cm, minimum height=0.6cm},
            activation/.style={draw, rounded corners, minimum height=1.2cm, minimum width=0.5cm, fill=white},
            model/.style={draw, rounded corners, minimum height=1.6cm, minimum width=5cm, fill=light-gray},
            arrow/.style={->,>=stealth},
        ]
            \node[text_node] (text_1) at (0,0) {I};
            \node[text_node] (text_2) at (1,0) {have};
            \node[text_node] (text_3) at (2,0) {a};
            \node[text_node] (text_4) at (3,0) {day};
            \node[text_node] (text_5) at (4,0) {off};
            \node[model] (model) at (2,1.5) {};
            \node[activation] (act_1) at (0,1.5) {};
            \node[activation] (act_2) at (1,1.5) {};
            \node[activation] (time_1) at (2,1.5) {};
            \node[activation] (time_2) at (3,1.5) {};
            \node[activation] (act_3) at (4,1.5) {};
            \draw[arrow] (text_1) -- ++(0, 0.7);
            \draw[arrow] (text_2) -- ++(0, 0.7);
            \draw[arrow] (text_3) -- ++(0, 0.7);
            \draw[arrow] (text_4) -- ++(0, 0.7);
            \draw[arrow] (text_5) -- ++(0, 0.7);
            \node[draw, rounded corners, minimum height=0.8cm, minimum width=1.5cm, fill=pink] (MLP) at (2.5,3.2) {Probe};
            \node[text_node] (text) at (2.5, 4.3cm) {DATE};
            \draw[arrow] (time_1) -- (MLP);
            \draw[arrow] (time_2) -- (MLP);
            \draw[arrow] (MLP) -- (text);
        \end{tikzpicture}
        \caption{Diagnostic probing (DP). It trains a probe on top of the contextual representations of the entity span (``a day'') to predict the label (``DATE'').}
        \label{fig:dp}
    \end{subfigure}\hfill
    \begin{subfigure}[t]{0.58\textwidth}
        \centering
        \begin{tikzpicture}[
            text_node/.style={draw, minimum width=0.8cm, minimum height=0.6cm},
            activation/.style={draw, rounded corners, minimum height=1.2cm, minimum width=0.5cm, fill=white},
            prefix_node/.style={draw, rounded corners, minimum height=0.5cm, minimum width=1.2cm, fill=pink, rotate=90},
            model/.style={draw, rounded corners, minimum height=1.6cm, minimum width=7cm, fill=light-gray},
            head/.style={draw, rounded corners, minimum height=0.8cm, minimum width=1.5cm, fill=light-gray},
            arrow/.style={->,>=stealth},
        ]
            \node[text_node] (text_1) at (1,0) {I};
            \node[] (text_2) at (2,0) {$\cdots$};
            \node[text_node] (text_4) at (3,0) {SEP};
            \node[text_node] (text_5) at (4,0) {a};
            \node[text_node] (text_6) at (5,0) {day};
            \node[text_node] (text_7) at (6,0) {EOS};
            \node[model] (model) at (3,1.5) {};
            \draw[arrow] (text_1) -- ++(0, 0.7);
            \draw[arrow] (text_2) -- ++(0, 0.7);
            \draw[arrow] (text_3) -- ++(0, 0.7);
            \draw[arrow] (text_4) -- ++(0, 0.7);
            \draw[arrow] (text_5) -- ++(0, 0.7);
            \draw[arrow] (text_6) -- ++(0, 0.7);
            \draw[arrow] (text_7) -- ++(0, 0.7);
            \node[prefix_node] () at (0,1.5) {Prefix};
            \node[activation] at (1,1.5) {};
            \node[] at (2,1.5) {$\cdots$};
            \node[activation] at (3,1.5) {};
            \node[activation] at (4,1.5) {};
            \node[activation] at (5,1.5) {};
            \node[activation] (last_activation) at (6,1.5) {};
            \node[head] (lm_head) at (6,3) {LM Head};
            \draw[arrow] (last_activation) -- (lm_head);
            \node[text_node] (text) at (6, 4cm) {DATE};
            \draw[arrow] (lm_head) -- (text);
        \end{tikzpicture}
        \caption{Probing via prompting (PP). We reformulate named entity labeling into an LM task by concatenating the given span (``a day'') with the sentence. We then use a prefix to instruct the model to predict the label (``DATE'').}
        \label{fig:pp}
    \end{subfigure}
    \caption{Illustration of different probing paradigms. Here, we show an example of named entity labeling: classifying a given entity span into pre-defined categories.}
    \label{fig:diagram}
\end{figure*}
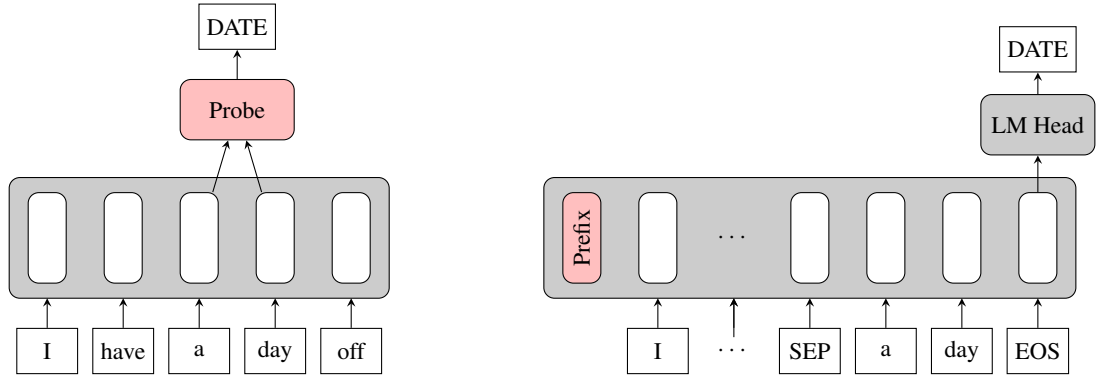
In this work, we address the above limitation with a novel probing framework that we call \defn{probing via prompting} (PP).
Drawing inspiration from recent work on prompting \cite{Larochelle2020, DBLP:journals/corr/abs-2107-13586}, we reformat a suite of probing tasks into question--answer pairs and instruct the model to answer the questions with a prefix \cite{li-liang-2021-prefix}. 
In effect, prompting acts as a model-free probe. %
Thus, the use of prompting instead of a diagnostic probe allows us to work around the dilemma of teasing apart what the representations contain versus what the probe learns.

In the empirical part of the paper, we conduct experiments on five linguistic tasks
and show that all these properties are indeed encoded
in the popular pre-trained language model, GPT-2. Furthermore, we show that probing via prompting appears to lead to insightful probing results.
In the language of \citet{hewitt-liang-2019-designing}, PP obtains high selectivity, i.e., the results show high accuracy on the target task, but, as expected, PP does not work well with random representations. 
These results suggest that PP learns little information on its own and the extracted linguistic properties are indeed encoded in the model.\looseness=-1

At a high level, this work takes the position that model-free probing methods such as PP are useful for accurately and faithfully locating and identifying
the linguistic properties embedded in these representations and helping us understand how neural language models
process text. 
In contrast to diagnostic probes, which require designing random model baselines and control tasks to control for the learning ability of the probe model, model-free probes like PP are less capable of learning the linguistic task themselves, and thus can naturally be more selective than diagnostic probes.

\section{Probing via Prompting}
We now introduce our probing via prompting framework (PP), illustrated in \cref{fig:pp}.

\begin{table*}
    \centering
    \begin{tabular}{@{}lcr@{}}
        \toprule
        Task & Context & Target Label \\
        \midrule
        POS & $\xx$ \sep \space brand \eos & NN \\
        Const. & $\xx$ \sep \space is a global brand \eos & VP \\
        Entity & $\xx$ \sep \space Disney \eos & ORG \\
        SRL & $\xx$ \sep \space is \sep \space The important thing about Disney \eos & ARG1 \\
        Coref. & $\xx$ \sep \space Disney \sep \space it \eos & True \\
        \bottomrule
    \end{tabular}
    \caption{Example prompt and target label for each task. $\xx=$``The important thing about Disney is that it is a global brand.'' The continuous prefix is neglected for simplicity.}
    \label{tab:ex}
\end{table*}

\subsection{Language Models}
Let $p$ be a language model with vocabulary $\Sigma$. 
In the case of an autoregressive language model,\footnote{Such language models are often called causal language models to differentiate them from cloze language models.} $p$ is defined as a distribution over $\Sigma^*$ that is locally normalized, i.e., for any $\boldsymbol{w} \in \Sigma^*$ we decompose $p(\boldsymbol{w})$ according to the chain rule as follows:
\begin{equation}
p(\boldsymbol{w}) = p(w_1) \cdot \prod_{i=2}^{|\boldsymbol{w}|} p(w_i \mid \boldsymbol{w}_{<i})
\end{equation}
Each ``local'' distribution $p(w_i \mid \boldsymbol{w}_{<i})$ is defined over $\Sigma$. 
Traditionally, language models include an $\eos$ symbol;
this means they produce a distribution over (an infinite number of) finite strings.

In this work, we focus on the case when $p$ is a Transformer-based language model \cite{NIPS2017_3f5ee243}---specifically, we take $p$ to be an instance of GPT-2 \cite{radford2019language}.
In contrast to most language models, GPT-2 dispenses with the $\eos$ symbol and therefore yields a distribution over infinite strings.\footnote{This yields a distribution over the $\omega$ language $\Sigma^{\omega}$.}
As it will be necessary for later discussion, we further introduce notation to discuss the internal workings of the Transformer.
We denote the layer activations $A^{(0)}, \ldots, A^{(L)}$, where $L$ is the total number of layers; the $0^{\text{th}}$ layer corresponds to the embedding layer. Here, $A^{(\ell)}=\left[\aaa_0^{(\ell)}, \ldots, \aaa_{|\ww|}^{(\ell)}\right]$ denotes the activation matrix of the $\ell^{\text{th}}$ layer and $\aaa_i^{(\ell)}$ is the activation vector for the token at position $i$.
The activation at the last layer is used to compute the distribution for the next token:
\begin{equation}
    p(w_{i+1} \mid \ww_{\leq i}) = \softmax(W\,\aaa_{i}^{(L)})
\end{equation}
where $W$ is a matrix that maps activations to logits over the vocabulary.\looseness=-1

\subsection{Edge Probing via Prompting}
The edge probing framework \cite{tenney-etal-2019-bert,tenney2018what} decomposes many common structured-prediction tasks into a common format of multi-class classification. 
In an edge probing task, a sentence $\boldsymbol{x}\in\Sigma^*$ and spans\footnote{Spans are contiguous substrings.} $\s_1,\s_2$ in $\xx$ are provided as inputs, and the goal is to select a correct label $y$ among a set of candidate labels $\mathcal{Y}$.
We have intentionally kept the definition of $\mathcal{Y}$ abstract because it is task-specific.
E.g., in named entity labeling, $\mathcal{Y}$ will be
a set of entity types, whereas in semantic role labeling $\mathcal{Y}$ will be a set of semantic roles.\footnote{The span $\s_2$ is to be omitted for single-span tasks such as entity labeling.}

Now we introduce how to perform edge probing via prompting. We follow the naming convention of \citet{schick-schutze-2021-exploiting} and begin describing our prompting approach by introducing a pattern--verbalizer pair.
\paragraph{Pattern.}
We convert $\xx, \s_1,\s_2$ into a string, called the \defn{pattern}, as follows
\begin{equation}
    \pp = \xx \circ \sep \circ \s_1 \circ \sep \circ \s_2 \circ \eos
\end{equation}
where $\circ$ is string concatenation.
Note that now $\pp\in\left(\Sigma\cup\{\sep,\eos\}\right)^*$.

\paragraph{Verbalizer.}
\newcommand{\verbalizer}{\mathtt{vb}}
Next, we define a \defn{verbalizer} function $\verbalizer :\mathcal{Y}\rightarrow \Sigma$ that maps each label to a token.\footnote{While one might think that an easy solution is to directly use category names as verbalizers, category names in edge probing tasks are often out-of-vocabulary words (e.g., \textsc{arg0}) that have to be decomposed into multiple sub-tokens. 
Thus, it is easier to simply introduce new symbols into the vocabulary for each class label.}
In our implementation, we introduce a distinguished symbol $\cls{y}$ for each label $y$.
Thus, our verbalizer becomes $\verbalizer:\mathcal{Y}\rightarrow \{\cls{1},\ldots,\cls{|\mathcal{Y}|}\}$,
where $|\mathcal{Y}|$ is the number of candidate classes.%

\paragraph{Inference.}
Now we augment the language model $p$ such that every conditional probability is over $\Sigma \cup \{\sep, \eos, \cls{1},\ldots,\cls{|\mathcal{Y}|}\}$ (instead of just $\Sigma$), so we expand $W$ %
correspondingly to incorporate the enlarged vocabulary. 
The newly added rows in $W$ and the embeddings of the newly added symbols are both randomly initialized from a normal distribution with a mean of $0$ and a variance of $0.02$ %
and not updated during training.
To make a prediction, we select the class whose verbalizer has the highest next-token probability:
\begin{equation}
    \widehat{y} = \argmax_{y' \in \mathcal{Y}} p \left( w_{|\pp|+1}=\verbalizer\left(y' \right) \mid \ww_{\leq |\pp|}\right)
\end{equation}
This completes our formalization of edge probing as prompting.

\subsection{Prefix Tuning}

To better instruct the language model to perform the target task, we prepend the pattern $\pp$ with a \defn{prefix} that is task-specific and independent of $\xx$, $\s_1$ and $\s_2$. Intuitively, we aim to steer a pre-trained language model to generate predictions using an instructive prefix.
For instance, a prefix for named entity labeling could be an additional string in $\Sigma^*$, e.g., ``classify the named entity into the following categories: person, organization \ldots''
However, in preliminary experiments, we found that discrete prefixes perform poorly on GPT-2---the prime object of our study in this paper.\footnote{This replicates the findings of \citet{li-liang-2021-prefix}, who also found that discrete prefixes performed poorly when applied to GPT-2 and BART; in their report, GPT-3 was the only exception.}
Thus, we resort to a continuous prefix \cite{li-liang-2021-prefix}.
The technical details of performing continuous prefix tuning in the case of a Transformer language model are given in \cref{app:prefix-tuning}.

\section{Experiments}
We empirically benchmark our pruning method against several previously proposed works.

\subsection{Tasks}
We experiment on five tasks derived from OntoNotes 5.0 \cite{AB2/MKJJ2R_2013}: part-of-speech tagging (POS), constituent labeling (const.), named entity labeling (entity), semantic role labeling (SRL), and coreference (Coref.). 
They are simplified versions of the original tasks in OntoNotes that are made compatible with the edge probing framework.
An example for each task is shown in \cref{tab:ex}.\looseness=-1

\subsection{Diagnostic Probing}
We compare PP with diagnostic probing (DP), and consider two types of diagnostic probes: a logistic regression probe (LR) and a multilayer perceptron probe (MLP).
The idea behind diagnostic probing is illustrated in \cref{fig:dp}.
In this paper, we train a DP to predict a label $y \in \mathcal{Y}$ for the given span(s) $\s_1$ (and optionally $\s_2$) of a sentence $\xx$, taking the contextual representations of the span(s) produced by the pre-trained model as inputs.

\begin{table}[t]
    \centering
    \begin{tabular}{@{}llcc@{}}
        \toprule
        Task & Method & Pre-trained & Random \\
        \midrule
        \midrule
        POS & PP & \textbf{94.28} & \textbf{13.14} \\
        & DP (MLP) & 94.01 & 47.89 \\
        & DP (LR) & 89.56 & 38.84 \\
        \midrule
        & Majority & \multicolumn{2}{c}{12.58}\\
        & Chance & \multicolumn{2}{c}{2.08} \\
        \midrule
        \midrule
        Const. & PP & \textbf{86.66} & \textbf{35.98} \\
        & DP (MLP) & 82.09 & 45.24 \\
        & DP (LR) & 71.32 & 42.67 \\
        \midrule
        & Majority & \multicolumn{2}{c}{35.66}\\
        & Chance & \multicolumn{2}{c}{3.33} \\
        \midrule
        \midrule
        Entity & PP & \textbf{93.81} & \textbf{15.91} \\
        & DP (MLP) & 88.43 & 35.87 \\
        & DP (LR) & 87.81 & 29.81 \\
        \midrule
        & Majority & \multicolumn{2}{c}{15.91}\\
        & Chance & \multicolumn{2}{c}{5.56} \\
        \midrule
        \midrule
        SRL & PP & \textbf{85.46} & \textbf{33.36} \\
        & DP (MLP) & 84.13 & 53.05 \\
        & DP (LR) & 77.43 & 47.99 \\
        \midrule
        & Majority & \multicolumn{2}{c}{33.36}\\
        & Chance & \multicolumn{2}{c}{1.52} \\
        \midrule
        \midrule
        Coref. & PP & \textbf{90.54} & \textbf{78.33} \\
        & DP (MLP) & 87.05 & \textbf{78.33} \\
        & DP (LR) & 81.21 & \textbf{78.33} \\
        \midrule
        & Majority & \multicolumn{2}{c}{78.33}\\
        & Chance & \multicolumn{2}{c}{50.00} \\
        \bottomrule
    \end{tabular}
    \caption{Accuracy ($\%$) for pre-trained GPT-2 (Pre-trained) and random GPT-2 (Random). }
    \label{tab:acc}
\end{table}

\subsubsection{Contextual Representations.}
DP takes a single vector as input, which represents the span(s) of interest under the context of the sentence. In this section, we introduce how to obtain such a vector from a pre-trained model.
For an input sentence $\xx$ of length $|\xx|$, we again denote the layer activations produced by the language model as $A^{(0)}, \ldots, A^{(L)}$ and $A^{(\ell)}=\left[\aaa_0^{(\ell)}, \ldots, \aaa_{|\xx|}^{(\ell)}\right]$.
Following \citet{peters-etal-2018-deep}, we pool the representations of the different layers into a \defn{scalar-mixed representation} as follows.
We define the matrix $A=\left[\aaa_0, \ldots, \aaa_{|\xx|}\right]$ with  $\aaa_i$ computed thusly:\looseness=-1
\begin{equation}
    \aaa_i = \sum_{\ell=1}^L n_{\textrm{MLP}}(\ell) \cdot \aaa^{(\ell)}_i
    \label{eq:scalar}
\end{equation}
where $n_{\textrm{MLP}}$ is a distribution over the layers $\{1,\ldots,L\}$ that is learned during training.\footnote{Note that we ignore the $0^{\text{th}}$ layer (embedding layer) for easier comparison in \cref{sec:pruning}.}
In case a span consists of multiple tokens, the per-token vectors (either the scalar-mixed vector $\aaa_i$ or the last layer activation $\aaa_i^{(L)}$) are combined into a span representation using self-attention pooling \cite{lee-etal-2017-end}. If more than one span exists, the span representations are concatenated into one.

\subsubsection{Baselines}
\paragraph{DP (LR).} 
The first diagnostic probe we consider is a multinomial logistic regression probe that resembles the classification head in \citet{cao-etal-2021-low}.
Following \citet{cao-etal-2021-low}, we compute the span representations using the activations $A^{(L)}$ from the last layer. The span representations are directly fed into a linear layer followed by a softmax output layer.\looseness=-1

\paragraph{DP (MLP).} 
The second diagnostic probe we consider is the MLP probe introduced by \citet{tenney2018what}.
Here, we use the scalar-mixed representations of $A$
to compute span representations, which are then fed into an MLP followed by a softmax output layer.\looseness=-1

\paragraph{Majority.}
Some tasks are highly imbalanced. For instance, over one third of the constituents (Const.) in the training set are adjective phrases (ADJP).
Therefore, for reference, we implement a majority baseline that always predicts the most frequent class.\looseness=-1

\subsection{Experimental Setup}
We investigate $\text{GPT2}_{\text{SMALL}}$ (which we refer to as simply GPT-2 in this paper), a Transformer
model with 12 layers and $117$M parameters pre-trained on a dataset of 8 million web pages \cite{radford2019language}. 
We also examine the probes on a random model with the same architecture as pre-trained GPT-2, but the parameters are randomly reset. Since the goal of probing is to inspect the information acquired during pre-training, an ideal probe should have low accuracy on the random model.\looseness=-1

For PP, we set the prefix length to 200 virtual tokens for tasks with unary edges (POS, Const., and entity) and 20 for those with binary edges (SRL, Coref.). 
For DP (MLP), we use a two-layer MLP with 512 hidden units. 
Following \citet{tenney2018what}, we linearly project
the per-token representations $\aaa_i$ into 256 dimensions before self-attentional pooling to improve performance. 
All models are trained for one epoch using the Adam optimizer \cite{DBLP:journals/corr/KingmaB14}. 
Our implementation is based on Hugging Face Transformers \cite{wolf-etal-2020-transformers}. Experiments are conducted on 8 Titan RTX GPUs.\looseness=-1

In our experiments, we only study English. Results may vary for other languages.
The English split contains 116K/16K/12K examples in the train/development/test sets, respectively. 
We train on the train set, experiment on the development set, and report final results on the test set.\looseness=-1

\subsection{Results}
\label{sec:results}
We compute the classification accuracy on each task and present the results in \cref{tab:acc}.
We observe that when GPT-2's parameters are randomly reset, %
PP performs substantially worse than the two diagnostic probes. 
Remarkably, the accuracy of PP only exceeds a majority-class classifier by a negligible amount on POS and Const., and is even identical to a majority-class classifier on entity, SRL and Coref. 
On the other hand, both DP (MLP) and DP (LR) outperform the majority-class baseline on all the tasks except for Coref., where the majority-class baseline already performed exceptionally well already. 
This result suggests that PP learns much less about the task on its own than DP, which makes it a better probe in terms of selectivity \citep{hewitt-liang-2019-designing}.\looseness=-1

\begin{table}
    \centering
    \begin{tabular}{@{}llccc@{}}
        \toprule
        Model & Probe & POS & POSC & $\Delta$ \\
        \midrule
        \midrule
        Pre-trained & PP & 94.28 & 74.48 & 19.80  \\
        & DP (MLP) & 94.01 & 69.58 & 24.43 \\
        & DP (LR) & 89.56 & 48.75 & 40.81 \\
        \midrule
        \midrule
        Random & PP & 13.14 & 7.66 &  5.48   \\
        & DP (MLP) & 47.89 &  45.71 & 2.18  \\
        & DP (LR) & 38.84 & 35.76 & 3.08  \\
        \midrule
        \midrule
        \multicolumn{2}{c}{Majority} & 12.58 & 6.58 & \\
        \multicolumn{2}{c}{Chance} & 2.08 & 2.08 & \\
        \bottomrule
    \end{tabular}
    \caption{POS and POSC accuracy of various methods on pre-trained GPT-2.}
    \label{tab:posc}
\end{table}

Meanwhile, when we consider pre-trained GPT-2, PP has higher accuracy on all the five probing tasks than DP (MLP) and DP (LR). 
We take these results to indicate that prompting works quite well at extracting linguistic knowledge. %
The fact that our more selective probe still performs well on linguistic tasks confirms that the considered linguistic information is indeed encoded in the pre-trained model.\looseness=-1%

\begin{figure*}[t]
    \centering
    \includegraphics{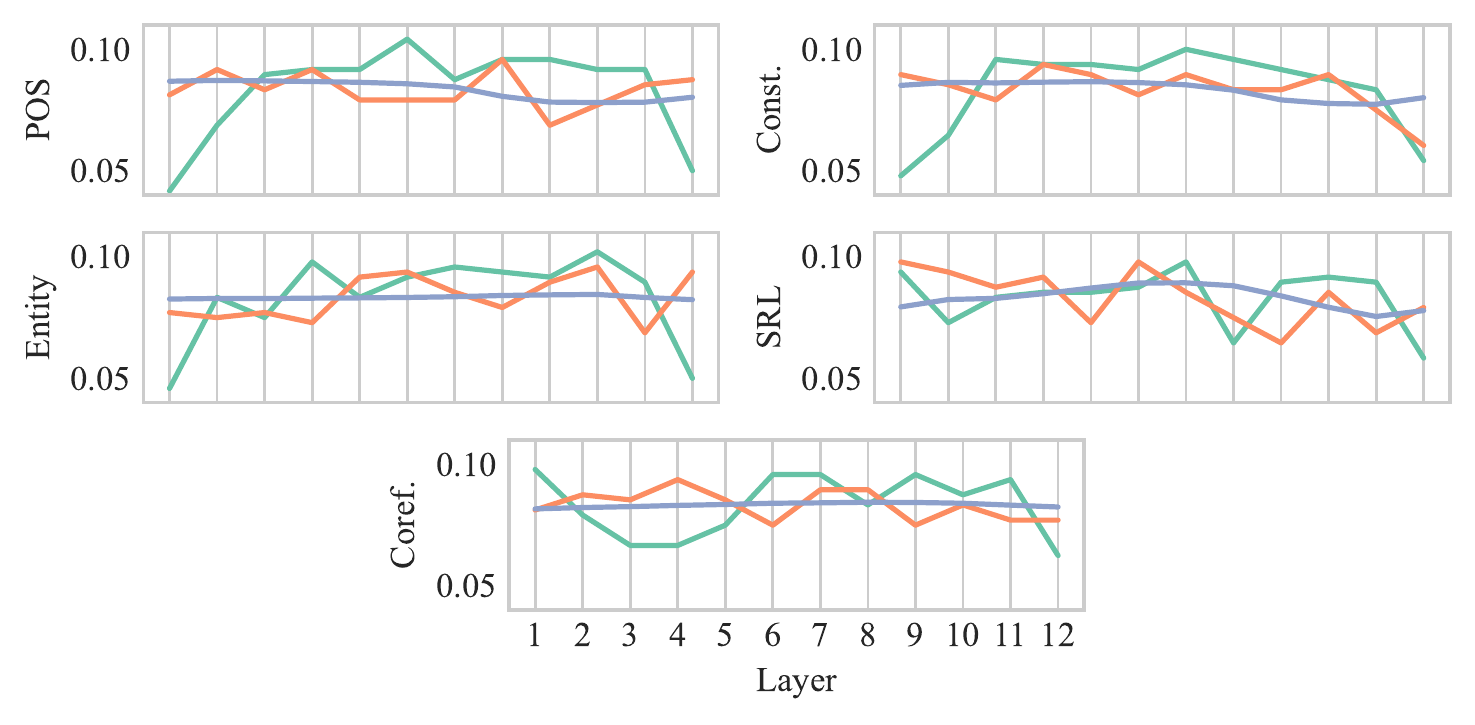}
    \caption{Layer distribution ($n$) for \textcolor{color1}{PP}, \textcolor{color2}{DP (MLP)}, and \textcolor{color3}{DP (LR)}.}
    \label{fig:layer}
\end{figure*}

\subsection{Control Tasks}
\citet{hewitt-liang-2019-designing} propose \defn{control tasks} to estimate a probe's selectivity in complement with random model baselines. A control task associates the inputs of a given linguistic task with random outputs. 
The key idea is that the control task can only be learned by the probe itself, so a selective probe should have high linguistic task accuracy but low control task accuracy. They further measure selectivity using a metric $\Delta$ as the difference between linguistic task accuracy and control task accuracy. 

In our experiments, we also create a control task, abbreviated POSC, for POS, where we randomly assign a POS tag for each distinct word. %
The results are shown in the first three rows in \cref{tab:posc}. To our surprise, we find that PP performs quite well on POSC, having an accuracy of $74.48\%$, which is only $19.80\%$ lower than its accuracy on POS. In contrast, the $\Delta$ metric for DP (MLP) and DP (LR) are $24.43\%$ and $40.81\%$ respectively. Therefore, if we were to use control tasks to measure selectivity, this result would suggest that PP is the least selective probe, which would be contradictory to our results in \cref{sec:results}, where we show the opposite.

To resolve the contradiction, we re-examine the implicit assumption behind control tasks: Randomness excludes the possibility of representations encoding the information of a control task, so that the control task accuracy can be solely attributed to the probe itself. If this were true, then the probe should be able to learn the task regardless of the representations it probes. To test that, we run the same experiments on the random model. The results are shown in rows 4--6 of \cref{tab:posc}. 
It is clear that accuracy on POSC under all three methods drops substantially when switching to the random model, which means the accuracy of a control task depends not only on the expressivity of the probe but essentially also on the representations.

\subsection{Discussion}
Since we found manually crafted prompts do not work well in our preliminary experiments, we resorted to prefix tuning. As a result, our prompting approach is not
fully parameter-free.
PP still involves learning parameters, and thus, we still run the risk of the prefix learning the target task on its own. Even though PP's performance on randomly initialized GPT-2 is barely better than that of the majority-class baseline, it is still much higher than chance, which indicates that PP still learns from the training set---for instance, it appears to learn the majority-class label. This is in line with the findings of \citet{zhong-etal-2021-factual} that an automatically optimized prompt can identify the majority-class label.\looseness=-1

Further study is needed to determine why PP performs worse on the random model but equally well or even better on the pre-trained model. 
PP is not simply less expressive because a less expressive model should perform worse on both pre-trained and random models, e.g., DP (LR) is less expressive than DP (MLP), but PP is the best on the pre-trained model and the worst on the random model. 
\citet{he2022towards} offer an interesting insight that continuous prompts and, in particular, prefix tuning can be regarded as adapters. 
Therefore, a possible explanation is that the adapter modules that are interlocked with the Transformer layer are more convoluted with the information encoded in the model than an external probe on top. When the model is pre-trained, they are able to apply modifications to the latent representations and steer the model on the fly to perform various tasks \cite{10.5555/3294771.3294820}, but if the model is randomly initialized, the noise hinders the learning of the task.

\section{Analysis}
Now that we have demonstrated the basic utility of PP, we attempt to use our methodology to determine where in the representations the information is encoded.
Thus, following \citet{cao-etal-2021-low}, we search for a subnetwork that optimizes the performance on the task of interest and analyze the resulting subnetwork. 
Since it has been shown that different attention heads in the Transformer capture different linguistic phenomena \cite{clark2019does, Voita:2019}, we prune attention heads instead of individual weights. 
Concretely, we use differentiable subset pruning (DSP) proposed by \citet{10.1162/tacl_a_00436}, which allows us to directly control the number of heads to keep. Pruning is performed jointly with prefix tuning.

\paragraph{Essential and Non-essential Heads.}
We call the $K$ heads that survive pruning \defn{essential heads} for the task, and those that are removed \defn{non-essential heads}. %
With $n_{\textrm{PP}}(\ell)$, we denote the distribution that is proportional to the number of essential heads in each layer $\ell$ under the PP scheme.
We define $n_{\textrm{LR}}(\ell)$ analogously.

\begin{figure}
    \centering
    \includegraphics{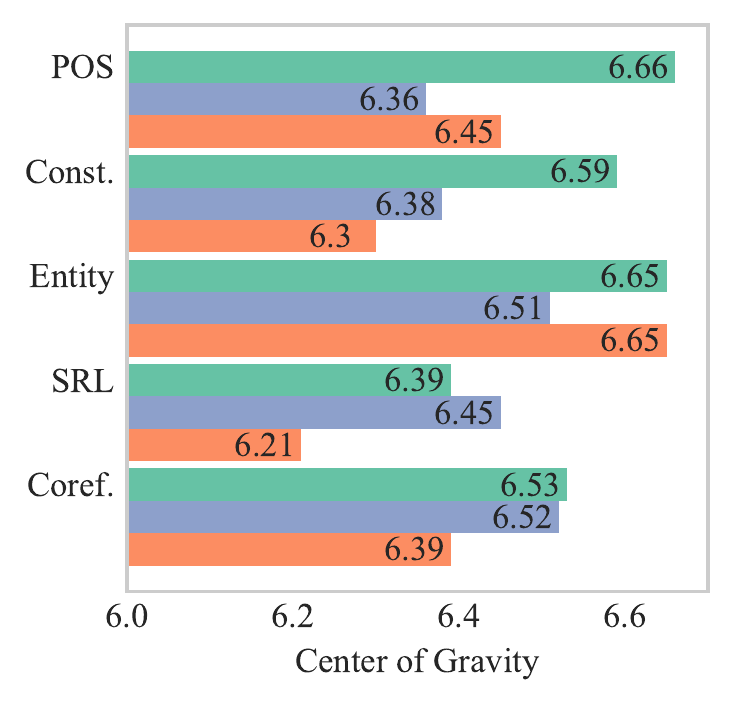}
    \caption{Center of gravity for \textcolor{color1}{PP}, \textcolor{color2}{DP (MLP)}, and \textcolor{color3}{DP (LR)}.}
    \label{fig:center}
\end{figure}

\subsection{Subnetwork Analysis}
\label{sec:pruning}

We now investigate how the essential heads of different tasks are distributed in the model.
To do so, we make use of the \defn{center of gravity} as a summary statistic introduced in \citep{tenney-etal-2019-bert}.
For any given layer distribution $n$, we compute the expected layer:
\begin{equation}
    \mathbb{E}_n[\ell] = \sum_{\ell=1}^L n(\ell) \cdot \ell
\end{equation}%
A higher center of gravity means the information for the task is encoded in higher layers.
In our experiments, we keep $K=96$ (out of $12\times12=144$) heads in GPT-2 and we report the average of 5 runs with different random seeds. \cref{fig:layer} depicts the layer distributions and \cref{fig:center} reports the centers of gravity.\looseness=-1

\citet{tenney-etal-2019-bert} find that BERT encodes linguistic knowledge in an order that follows the classical NLP pipeline: syntactic information is stored in earlier layers than semantic information. 
As shown in \cref{fig:center}, we are able to reproduce their results on GPT-2 using DP (MLP). Specifically, the tasks are encoded from the bottom of the model to the top in the following order: POS $\rightarrow$ Const. $\rightarrow$ SRL $\rightarrow$ entity $\rightarrow$ Coref.  \citet{cao-etal-2021-low} also find that entity is localized in higher layers than POS. We obtain the same results using DP (LR). 

However, the other three tasks (SRL, Const., Coref.) all have lower centers of gravity than POS, which contradicts the order of \citet{tenney-etal-2019-bert} as POS is believed to be the most basic syntactic information and should appear the earliest.
Moreover, PP produces an order that is entirely different from both DP methods: SRL $\rightarrow$ Coref. $\rightarrow$ Const. $\rightarrow$ entity $\rightarrow$ POS. Noticeably, according to PP, syntactic information (POS and Const.) is captured in higher layers on average than what is discovered by DP (MLP). This is in agreement with findings from recent unsupervised probing works \cite{gupta:2020,zhou:21}.

In conclusion, we find that different probing and analysis methods can lead to drastically different results.
Since the choice of probing methodology influences the resulting ordering, we believe that future work should be cautious in making claims based on a single interpretation approach.
Instead, a number of probing methods should be considered.
\looseness-1

\begin{table}
    \centering
    \begin{tabular}{@{}lccc@{}}
        \toprule
         & Essential & Non-essential & Majority \\
        \midrule
        POS & 93.21 & 1.9 & 12.58 \\
        Const. & 84.61 & 7.66 & 35.66 \\
        Entity & 90.00 & 4.50 & 15.91 \\
        SRL & 70.34 & 1.74 & 33.36 \\
        Coref. & 85.50 & 58.14 & 78.33 \\
        \bottomrule
    \end{tabular}
    \caption{Accuracy ($\%$) of PPP models with only essential heads or non-essential heads.}
    \label{tab:downstream}
\end{table}

\subsection{Amnesic Probing}
\citet{ravichander-etal-2021-probing} and \citet{Lasri2022ProbingFT} argue that a high-accuracy probe does not necessarily mean the information is important or used by the model.
For instance, linguistic properties can be spuriously encoded. To investigate whether a property is actually used by the model in prediction, \citet{10.1162/tacl_a_00359} propose amnesic probing, which neutralizes some information from the representation and measures the influence of that intervention. 
In the same spirit, we remove the information of a given property by discarding the essential heads from GPT-2, evaluate the pruned model on the WikiText-103 LM dataset \cite{merity2016pointer}, and quantify the importance of that property with the absolute increase in cross-entropy.
By keeping the number of pruned heads constant, we control for the amount of information removed on each task.
Note that the performance degradation of the LM cannot be solely attributed to the inspected property, as additional confounding information may also be removed when we discard essential heads for a property. 
Yet, it is still an indicator of the relative importance of different properties.\looseness=-1

In order to make sure the information for the targeted properties is eliminated from the model, we first evaluate PP models with only non-essential heads on the linguistic tasks. We keep $144-96=48$ non-essential heads in the model. 
Again, the average of five runs with different random seeds is reported. \cref{tab:downstream} shows that the models with only non-essential heads perform substantially worse than the models with essential heads and even the majority baseline, which shows that the model has lost its ability to predict a property after the target property's essential heads are removed. 
Next, we inspect how much impact it would have on the pre-training task.
The results of LM loss are summarized in \cref{tab:amnesic}.
For reference, we also evaluate the model with $48$ random heads (Random). 
Generally, all five linguistic properties are useful for LM, as leaving out their essential heads all lead to a bigger increase in LM loss than Random.%
Entity is clearly the most important property, as removing its essential heads leads to an increase of $4.22$ in LM loss. Coref. is the second, accounting for $3.97$ loss increase. POS and Const. are almost equally important. SRL is the least important factor, causing only $0.1$ more damage than Random. 
Our results demonstrate that probing accuracy in \cref{tab:acc} (POS $>$ entity $>$ Coref. $>$ Const. $>$ SRL) is not reflective of the property's importance according to \cref{tab:amnesic} (entity $>$ Coref. $>$ POS $>$ Const. $>$ SRL), which is consistent with the findings of \citet{10.1162/tacl_a_00359}.\looseness=-1

\section{Related Work}
\paragraph{Probing.} 
There has been a plethora of research papers analyzing the neural representations of NLP models. 
One of the primary goals of such research is to understand whether the linguistic information commonly believed to be important for representing language is actually captured in the representations. The most popular approach for associating network components with linguistic properties is to train an auxiliary model to predict such properties from activations of neural networks \cite{10.1162/tacl_a_00254}. This technique is now commonly referred to as probing \cite{conneau-etal-2018-cram, alain2018understanding, saleh-etal-2020-probing, tenney2018what}, but has also been known as auxiliary prediction \cite{DBLP:journals/corr/AdiKBLG16, zhang-bowman-2018-language}, diagnostic classification \cite{Veldhoen2016DiagnosticCR, ijcai2018-796, giulianelli-etal-2018-hood}, and others \cite{belinkov-etal-2017-evaluating, peters-etal-2018-dissecting, naik-etal-2018-stress}. However, the interpretation of probing results has been called into question \cite{hewitt-liang-2019-designing}: Do the representations encode the linguistic information or does the probe learn the task on its own? A commonly held belief \cite{alain2018understanding, liu-etal-2019-linguistic, hewitt-manning-2019-structural} is a simple model (e.g. a linear one) is not capable of learning the task itself and thus is preferred, but \citet{pimentel-etal-2020-information} argues that one should always choose the best possible probe because it reveals the most linguistic information present in the representations. \citet{pimentel-etal-2020-pareto, voita-titov-2020-information} explicitly model the trade-off between accuracy and model complexity. \citet{cao-etal-2021-low} propose to search for a subnetwork within the model rather than train an auxiliary model, but a task-specific classification head is still required. 

\begin{table}
    \centering
    \begin{tabular}{@{}lcc@{}}
        \toprule
         & LM Loss & $\Delta$\\
        \midrule
        Vanilla & 3.42 & --- \\
        \midrule
        Random & 6.94 & 3.52 \\
        \midrule
        POS & 7.21 & 3.79 \\
        Const. & 7.17 & 3.75 \\
        Entity & 7.64 & 4.22 \\
        SRL & 7.04 & 3.62 \\
        Coref. & 7.39 & 3.97 \\
        \bottomrule
    \end{tabular}
    \caption{LM loss on WikiText-103 of vanilla GPT-2 (Vanilla), GPT-2 whose heads are randomly removed (Random), and GPT-2 whose essential heads for different properties are removed.}
    \label{tab:amnesic}
\end{table}
\paragraph{Prompting.} 
The deep contextual word representations are typically derived from either an LM \cite{peters-etal-2018-deep, Radford2018ImprovingLU} or a masked LM \cite{Devlin:2018}.
A common use of these pre-trained language models is fine-tuning. However, an alternative approach called prompting has recently gained much popularity. Instead of accommodating a language model for downstream tasks, prompting adapts downstream tasks to be more like LM with the aid of a prompt. In this way, the pre-trained model can be used to perform few-shot or even zero-shot learning \cite{petroni-etal-2019-language, Larochelle2020, JMLR:v21:20-074, schick-schutze-2021-exploiting, schick-schutze-2021-just}. 
Most papers construct templates with blanks for the model to fill. For example, LAMA \citep{petroni-etal-2019-language} creates cloze-style templates to probe knowledge; \citet{Larochelle2020} put task descriptions and examples in the prefix and then the model performs various tasks by finishing the sentence; \citet{cui-etal-2021-template} enumerate every possible text span in a sentence, create a template for each of them, and fine-tune the model to perform named entity recognition (NER). However, creating such templates requires a large amount of time and human expertise, and does not necessarily do well on the task of interest. Therefore, many researchers focus on generating prompts automatically \cite{10.1162/tacl_a_00324, shin-etal-2020-autoprompt, haviv-etal-2021-bertese, gao-etal-2021-making}. The prompts must not consist of natural language, but can also be continuous vectors \cite{li-liang-2021-prefix, qin-eisner-2021-learning, lester-etal-2021-power, hambardzumyan-etal-2021-warp}. We refer the reader to \citet{DBLP:journals/corr/abs-2107-13586} for a more thorough survey about prompting. In this work, we apply the method of \cite{li-liang-2021-prefix} to learn continuous prompts to instruct the model to predict linguistic structure.

\paragraph{Pruning.} Neural network pruning aims to reduce the model size and increase inference speed by removing redundant network components, such as individual parameters \cite{lecun, Hassibi:93, han2015learning}, convolutional channels \cite{liu2017learning, Luo_2017_ICCV, he2017channel}, and attention heads \cite{Michel:2019, Voita:2019, 10.1162/tacl_a_00436}. In addition to model compression, pruning has also been used for analysis: \citet{Voita:2019} analyze the linguistic roles the unpruned heads play; \citet{cao-etal-2021-low} look at the location of unpruned weights. Similarly, we examine the network components that survive pruning, but we apply head pruning \cite{10.1162/tacl_a_00436} instead of weight pruning \cite{louizos2017learning} since attention heads are believed to be more linguistically interpretable than weights.

\section{Conclusion}
With the growing popularity of probing, there have been increasing concerns that high probing performance on a linguistic property cannot be attributed to representations encoding the property, since the probe can learn the probing task on its own. In this work, we propose a novel probing via prompting method, which drastically reduces the probe's ability to learn and, thus, mitigates this problem. 

We conduct experiments on five linguistic tasks and show that these properties are indeed encoded in one popular pre-trained language model, GPT\nobreakdash-2. However, they might not be encoded in a natural progression in the model as previously believed. For further study, we hope to develop tools that can more accurately and faithfully locate and identify the linguistic properties embedded in the model and help us understand the way neural language models process text. 

\section*{Ethical Considerations}
The OntoNotes 5.0 dataset \cite{AB2/MKJJ2R_2013}, licensed through LDC, annotates various genres of texts in three languages (English, Chinese, and Arabic) with structural information and shallow semantics. 
OntoNotes 5.0 inevitably contains personal information and offensive content. 
However, we only run experiments on the dataset and do not disseminate it or our trained models, which are only available upon request.
We also make sure the examples shown in our paper are anonymized. 
The pre-trained language model GPT-2 can also encode certain social biases \cite{liang2021towards}. Our research in probing could help us understand and mitigate these biases.

\section*{Acknowledgements}
This publication was made possible by an ETH AI Center doctoral fellowship to Jiaoda Li.
Ryan Cotterell acknowledges support from the Swiss National Science Foundation (SNSF) as part of the ``The Forgotten Role of Inductive Bias in Interpretability'' project.

\bibliography{anthology,custom}
\bibliographystyle{acl_natbib}

\appendix
\onecolumn
\section{Prefix Tuning}\label{app:prefix-tuning}
\subsection{Background: Transformer}
Recall that in a Transformer-based causal language model \cite{NIPS2017_3f5ee243, radford2019language}, each layer consists of two sub-layers: a multi-head self-attention mechanism and a fully connected feed-forward network. Each sub-layer is short-circuited with a residual connection, followed by layer normalization. 
Now we zoom in on the self-attention mechanism. For the sake of illustration, we assume there is only one head in each sub-layer. 
In self-attention, each activation vector $\aaa_i^{(\ell)}$ is first linearly projected into three vectors: query $\qq_i^{(\ell)}=W_q^{(\ell)} \aaa_i^{(\ell)}\in\R^d$, key $\kk_i^{(\ell)}=W_k^{(\ell)} \aaa_i^{(\ell)}\in\R^d$, and value $\vv_i^{(\ell)}=W_v^{(\ell)} \aaa_i^{(\ell)}\in\R^d$. Then we compute the output $\zz_i^{(\ell)}$ as the sum of values weighted by a compatibility score between query and key.
\begin{equation}
\label{eq:att}
    \zz_i^{(\ell)} = \sum_{j=0}^i \softmax\left(    \frac{{\qq_i^{(\ell)}}^\top \kk_j^{(\ell)}}{\sqrt{d}}\right)_j \vv_j^{(\ell)}
\end{equation}
The upper bound $i$ in the summand makes sure it can not attend to subsequent positions, and thereby the prediction for the next token at position $i+1$ can only depend on the tokens at positions up to $i$.
We abstract the feed-forward sub-layer, residual connections, and layer normalizations with a function $f$ and so we have
\begin{equation}
    \aaa_i^{(\ell+1)} = f\left(\aaa_i^{(\ell)}, \zz_i^{(\ell)} \right)
\end{equation}

\subsection{Prefix Tuning}
Prefix tuning prepends the pattern $\pp$ a prefix of length $T$, which we index from $-T$ to $-1$. Then \cref{eq:att} becomes
\begin{equation}
    \widetilde{\zz}_i^{(\ell)} = \sum_{j=-T}^i \softmax\left(    \frac{{\qq_i^{(\ell)}}^\top \kk_j^{(\ell)}}{\sqrt{d}}\right)_j \vv_j^{(\ell)}
\end{equation}
and the layer activations are modified accordingly:
\begin{equation}
    \widetilde{\aaa}_i^{(\ell+1)} = f\left(\widetilde{\aaa}_i^{(\ell)}, \widetilde{\zz}_i^{(\ell)} \right)
\end{equation}
The $0^{\text{th}}$ layer is left unchanged: $\widetilde{\aaa}_i^{(0)}=\aaa_i^{(0)}$.
Note that we never compute activations $\aaa_i^{(\ell)}$ for the prefix ($i<0$), so we do not need queries $\qq_i^{(\ell)}$ for them, and the key--value pairs $\kk_i^{(\ell)}, \vv_i^{(\ell)}$ cannot be obtained through projection from $\aaa_i^{(\ell)}$. Instead, they are learned directly. 
During training, only $\kk_i^{(\ell)}, \vv_i^{(\ell)}$ for prefix are learned while the parameters of the language model are frozen. During inference, the modified activations are now used:
\begin{equation}
    \widetilde{p}(w_{|\pp|+1} \mid \ww_{<|\pp|+1}) = \softmax(W \widetilde{\aaa}_{|\pp|}^{(L)})
\end{equation}
\begin{equation}
    \widehat{y} = \argmax_{y' \in \mathcal{Y}} \widetilde{p} \left( \ww_{|\pp|+1}=\verbalizer\left(y' \right) \mid \ww_{<|\pp|}\right)
\end{equation}

\end{document}